\documentclass[fleqn,10pt]{wlscirep}
\usepackage[utf8]{inputenc}
\usepackage[T1]{fontenc}
\usepackage{graphicx}
\usepackage{subcaption}
\usepackage{lipsum}
\usepackage{array, makecell}
\usepackage{outlines}

\usepackage{multirow}
\usepackage{ragged2e}
\newcolumntype{C}[1]{>{\centering\arraybackslash}p{#1}}

\usepackage[section]{placeins}

\title{Deep learning-based identification of patients at increased risk of cancer using routine laboratory markers}

\author[1,*]{Vivek Singh}
\author[1]{Shikha Chaganti}
\author[2]{Matthias Siebert} 
\author[1]{Sowmya Rajesh}
\author[3,4]{Andrei Puiu}
\author[5]{Raj Gopalan} 
\author[6]{Jamie Gramz} 
\author[1]{Dorin Comaniciu} 
\author[1]{Ali Kamen}
\affil[1]{Siemens Healthineers, Digital Technology and Innovation, Princeton, 08540, USA}
\affil[2]{Siemens Healthineers, Digital Technology and Innovation, Erlangen, 91052, Germany}
\affil[3]{Siemens SRL, Advanta, Brasov, 500007, Romania}
\affil[4]{Transylvania University of Brasov, Automation and Information Technology, Brasov, 500174, Romania}
\affil[5]{Siemens Healthineers, Laboratory Diagnostics, Tarrytown, NY 10591, USA.}
\affil[6]{Siemens Healthineers, Digital and Automation, Malvern, PA 19355, USA.}
\affil[*]{vivek-singh@siemens-healthineers.com}

\keywords{Oncology Screening, Deep Learning, Laboratory Markers}

\begin{abstract}

Early screening for cancer has proven to improve the survival rate and spare patients from intensive and costly treatments due to late diagnosis. Cancer screening in the healthy population involves an initial risk stratification step to determine the screening method and frequency, primarily to optimize resource allocation by targeting screening towards individuals who draw most benefit. For most screening programs, age and clinical risk factors such as family history are part of the initial risk stratification algorithm. In this paper, we focus on developing a blood marker-based risk stratification approach, which could be used to identify patients with elevated cancer risk to be encouraged for taking a diagnostic test or participate in a screening program. We demonstrate that the combination of simple, widely available blood tests, such as complete blood count and complete metabolic panel, could potentially be used to identify patients at risk for colorectal, liver, and lung cancers with areas under the ROC curve of 0.76, 0.85, 0.78, respectively. Furthermore, we hypothesize that such an approach could not only be used as pre-screening risk assessment for individuals but also as population health management tool, for example to better interrogate the cancer risk in certain sub-populations.

\end{abstract}
\begin{document}

\flushbottom
\maketitle
\thispagestyle{empty}

% \noindent Please note: Abbreviations should be introduced at the first mention in the main text – no abbreviations lists. 

\section*{Introduction}
This paper focuses on the use of multiple biomarkers for the assessment of patients, or identification of otherwise healthy individuals, who are at increased risk of cancer. With the high mortality rate associated with cancer patients, significant research has been conducted to help identify patients at higher risk, starting with identifying medical conditions that increase the risk of cancer, such as diabetes, or genetic predispositions that promote its development\cite{Samadder2021}. Furthermore, various screening procedures have been developed to help facilitate early diagnosis such as the Faecal Immunochemical Test (FIT) and colonoscopy for colorectal cancer (CRC)\cite{Rex2017}, mammography for breast cancer\cite{Mendes2021}, and low-dose computed tomography (LDCT) for lung cancer\cite{deKoning2020}. However, cancer screening rates and their uptake remains lower than desired, e.g., in the US\cite{Hall2018}. While there are several factors contributing to this low uptake, one of the key factors is the lack of awareness within the general population. This is even more important to address for people who may be at increased risk and would benefit from early and/or regular screening. In other words, there is still a need for convenient tests for early detection of rapidly progressing diseases such as cancer so that intervention can start as early as possible\cite{Philipson2023}. 
% This has also been concluded by studies that have been conducted using the existing cancer screening tools.

%Due to high prevalence and high mortality associated with cancer, 
Several cancer risk prediction/assessment tools based on demographic, socioeconomic or blood based markers have been developed over the years, and studies have shown that cancer risk assessment algorithms could have an impact in early cancer diagnosis\cite{Kostopoulou2022}. For instance, the Qcancer 10 year risk algorithm\cite{HippisleyCox2015} considers the age, ethnicity, deprivation, body mass index, smoking, alcohol, previous cancer diagnoses, family history of cancer, relevant comorbidities, and medication data for a patient and predicts the cancer risk for 11 types of cancers. Nartowt et al.\cite{Nartowt2020} reported high concordance in the prediction of CRC into low, medium and high groups using an artificial neural network trained on patient data comprising age, sex, and complete blood count (CBC). ColonFlag\cite{Goshen2018} can be used to identify individuals at high risk of CRC using specific blood-based markers and refer them to screening procedures such as colonoscopy. More recently, a cell-free DNA-based blood test for the early detection of CRC has been clinically validated in the ECLIPSE study\cite{Chung2024}. Moreover, multi-cancer early detection technologies\cite{Hackshaw2022} such as the Galleri test\cite{Klein2021, Shao2022} can identify abnormal methylation patterns in cell-free DNA to detect a cancer signal and predict its origin.

Besides algorithm development, the deployment of the algorithm and communication of the findings play a critical role in acceptance and clinical use of the algorithm\cite{Hamilton2009}, and must be taken into account to facilitate screening uptake. To this end, instead of defining a test with a specific set of ingredients catered towards a particular cancer, we propose to use commonly measured blood markers, often obtained during the annual physical exam, and obtain a risk profile for multiple cancers. Furthermore, instead of reporting a risk score, we compute the pre- and post-test odds of a patient at risk of developing cancer over the next 12 months.

A key challenge in developing a model that considers several biomarkers is to deal with a significant degree of missingness in the historical data as not all markers may be obtained at each encounter. Although this issue can be partly alleviated by considering the biomarkers obtained at an annual physical exam, we observed that in real world data, there is still a significant degree of missingness, either due to a lack of awareness, insurance coverage or reimbursement, among other reasons. A standard approach to deal with missingness in input data is to impute the missing values using statistical methods, such as expectation maximization and regression\cite{Heymans2022}. However, the quality of imputed data is limited and can significantly impact the generalization ability of the trained model.

\begin{figure}[ht]
    \centering
    \includegraphics[width=0.9\linewidth]{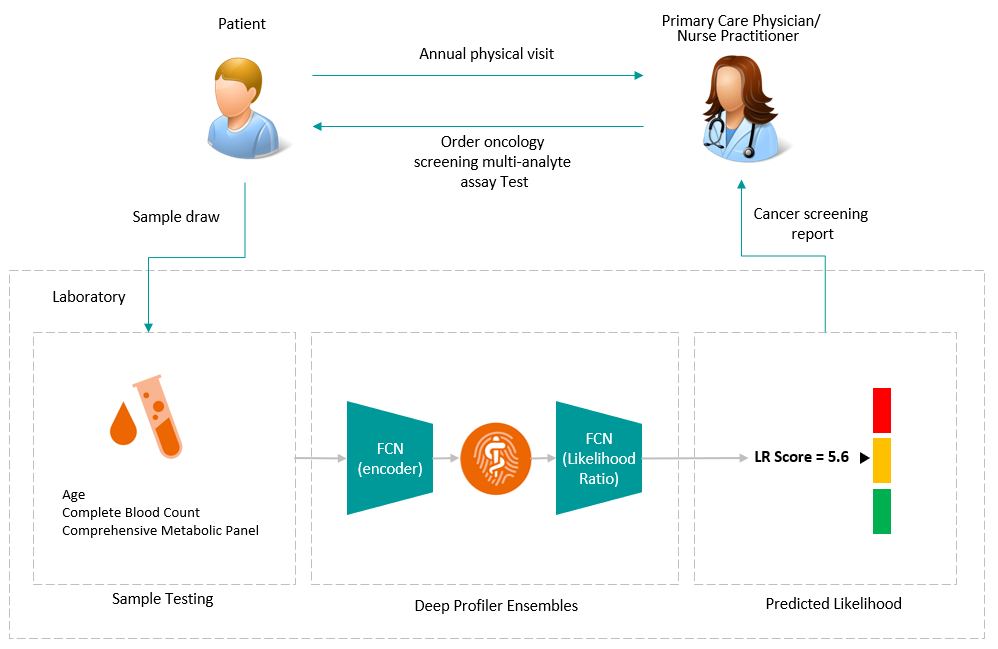}
    \caption{Workflow of using a biomarker-based pre-screening test.}
    \label{fig:grahicalabstract}
\end{figure}

In this work, we address the aforementioned challenges by training a deep learning model, \emph{Deep Profiler}, which takes the age, sex, and commonly obtained blood biomarkers included in CBC and Comprehensive Metabolic Panel (CMP), and outputs a likelihood ratio of a patient to develop cancer over the period of the following 12 months (see Figure \ref{fig:grahicalabstract}). The Deep Profiler architecture employs a variational autoencoder (VAE) model that is pre-trained to impute missing data similar to the masked language modeling technique. Subsequently, we train cancer-specific risk prediction models from the shared encoded latent space and compute the likelihood ratio for each patient. We validate the proposed method over screening-relevant cohorts for three different cancers - colorectal, liver, and lung. These are among the top cancers responsible for cancer related mortality rate in the US (https://seer.cancer.gov/statfacts/html/common.html, accessed April 30, 2024.).
%In summary, contributions from this paper are 3-fold: firstly, we present an approach to identify patients at increased risk of developing cancer only using routine blood markers obtained during annual physical exams; secondly, we use SHAP analysis using likelihood ratio, instead of model's output scores, which allows identification of relevant features while taking model uncertainty into account, and finally, we characterise the models with respect to comorbidities as potential confounders to better identify patient sub-groups for whom the model could be most appropriate to use.

\section*{Results}
\subsection*{Patient characteristics}
To evaluate Deep Profiler across multiple cancer types, we created a cohort of individuals that either have no diagnosis of cancer, or were diagnosed with malignant neoplasms of either colorectal, liver or lung. The validation cohorts each include nearly 5k patients with no cancer diagnosis, and 94, 189 and 224 patients who develop colorectal, liver or lung cancer, respectively. Note that the patients in the validation set were not used during model development and training. The cohorts used for model development include nearly 10k patients with no cancer diagnosis, and 293, 626 and 683 patients who develop colorectal, liver or lung cancer, respectively. 
Supplementary Table S1 summarizes the statistics of various biomarkers, including age and sex, over the entire cohort covering all three cancer types. Significant differences in the distributions of biomarker measurements in cancer cohorts as compared to control cohorts are indicated.

\subsection*{Likelihood ratio}
\begin{figure}[ht]
    \centering
    \begin{subfigure}[t]{\linewidth}
        \includegraphics[width=\linewidth]{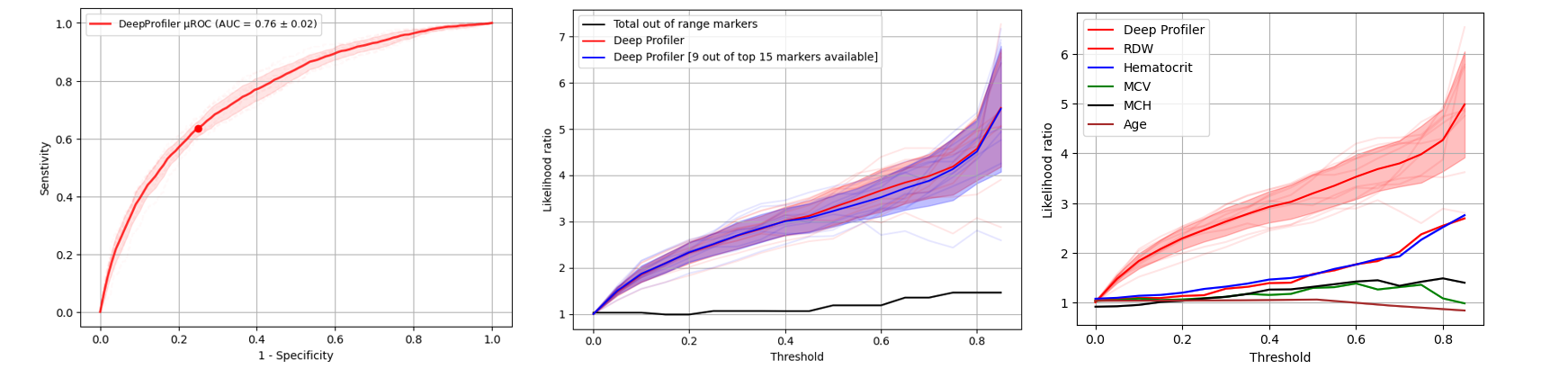}
        \caption{Colorectal cancer}\label{fig:crc_performance}
    \end{subfigure}
    \begin{subfigure}[b]{\linewidth}
        \includegraphics[width=\linewidth]{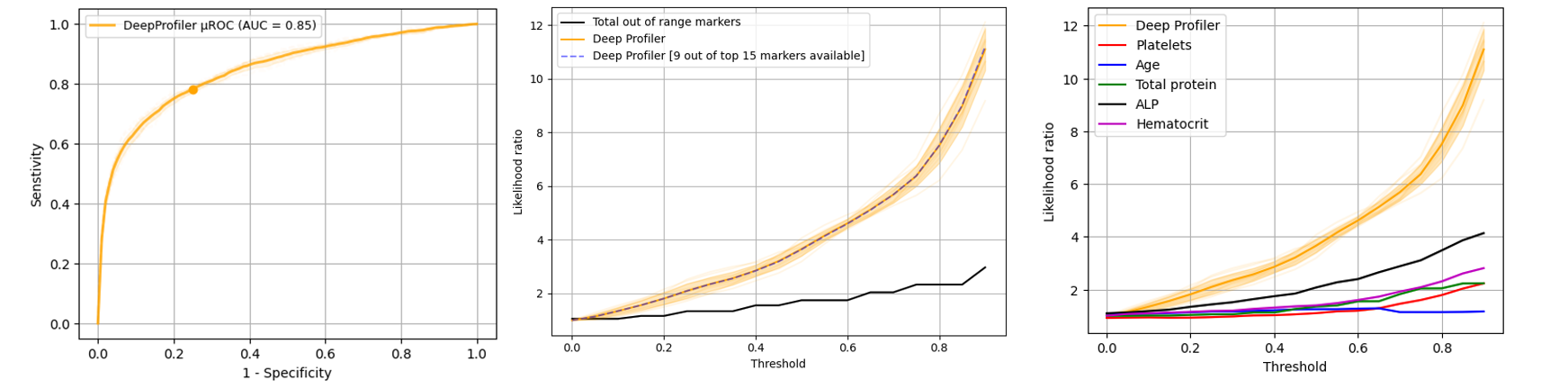}
        \caption{Liver cancer}\label{fig:liver_performance}
    \end{subfigure}
    \begin{subfigure}[b]{\linewidth}
        \includegraphics[width=\linewidth]{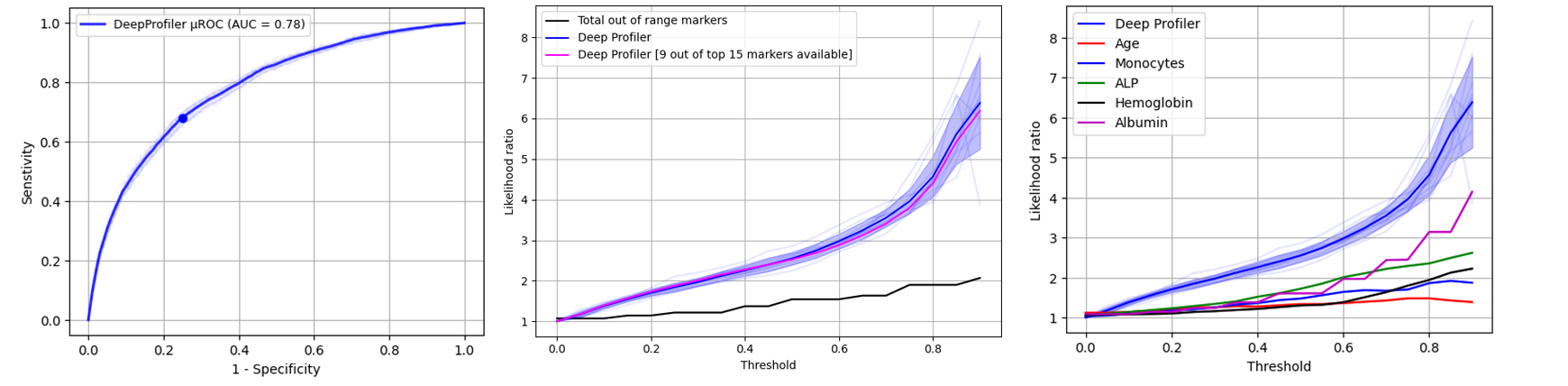}
        \caption{Lung cancer}\label{fig:lung_performance}
    \end{subfigure}
    \caption{Quantitative performance assessment on (a) colorectal, (b) liver, and (c) lung cancer validation cohorts. ROC curves represent the predictions of the corresponding models (left). %, with optimal operating points indicated. 
    Likelihood ratio plots show the likelihood ratios of corresponding models on the full cohort and the subgroup of patients having at least nine out of the top 15 markers available, for increasing risk thresholds (middle). Likelihood ratio curves are compared to those of baseline models that are either based on the total number of OoR markers (middle) or single markers (right), corresponding to the top five markers for each cancer type (cf. Figure \ref{fig:cohort_shap}), including age for CRC. Thin lines depict likelihood ratio curves of base models, while ribbons correspond to one standard deviation of their likelihood ratios. ALP, alkaline phosphatase; MCH, mean corpuscular hemoglobin; MCV, mean corpuscular volume; RDW, red cell distribution width.}
    \label{fig:performance}
\end{figure}

We train Deep Profiler ensembles to estimate cancer risk for three cancer types: colorectal, liver and lung cancer. Given the biomarker values, the model outputs an array of risk scores (from the ensemble) for each cancer type, which are then utilized to compute the likelihood ratio (LR) of post- to pre-test odds. LRs at increasing risk thresholds, together with Receiver Operating Characteristic (ROC) curves, are shown in Figures \ref{fig:crc_performance}, \ref{fig:liver_performance}, and \ref{fig:lung_performance} for colorectal, liver, and lung cancer, respectively (corresponding precision-recall curves are provided in Supplementary Figure S1).

Since N{\ae}ser et al. reported that the probability of cancer increased with the number of test results outside the reference range\cite{Naeser2017}, we used the total out-of-reference-range (OoR) markers as a baseline. Indeed, cancer likelihood increases with the total OoR markers. However, the increase is significantly lower as compared to the Deep Profiler models. Similarly, when compared individually with any of the top-5 markers identified by the model for each cancer, Deep Profiler provides a 2-3 fold improvement.

In CRC, age is the only factor considered to determine screening eligibility according to recommendations by the US Preventive Services Task Force (USPSTF)\cite{USPSTF2021_CRC}. Therefore, we also provide an LR curve for age (normalized between 40 and 85 years) as the single indicator. While CRC shows increased prevalence until the age of about 60, the increase of LR provided by Deep Profiler is significantly higher, with a 4-fold improvement in LR at a threshold greater than 0.8.

Overall, the liver-specific Deep Profiler model performs significantly better as compared to our Deep Profiler models for CRC and lung cancer (LR of \textasciitilde{7.5} vs. \textasciitilde{4.5} at a threshold of 0.8, respectively), which is also reflected in the corresponding ROC curves. Importantly, performance is preserved when assessed in cohort-specific subsets that only include cases for which measurements for at least nine of top-15 markers are available, indicating that the models did not learn patterns related to missing marker measurements.

\subsection*{Relevant biomarkers by cancer type}
\begin{figure}[ht]
    \setbox1 = \hbox{\includegraphics[keepaspectratio=true]{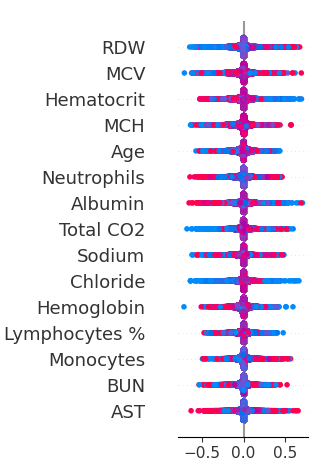}}
    \setbox2 = \hbox{\includegraphics[keepaspectratio=true]{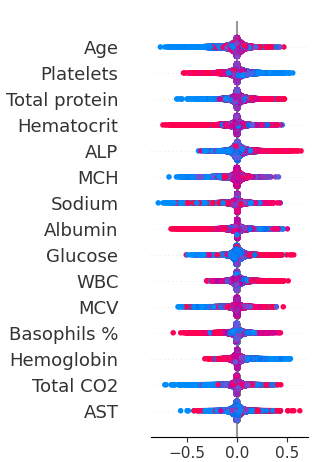}}
    \setbox3 = \hbox{\includegraphics[keepaspectratio=true]{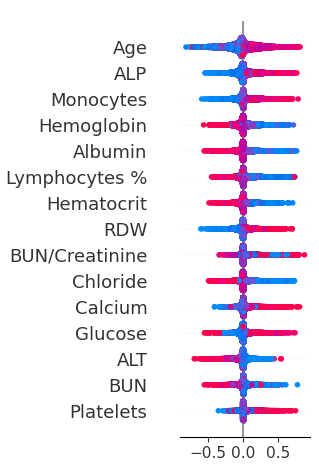}}
    \setbox4 = \hbox{\includegraphics[keepaspectratio=true]{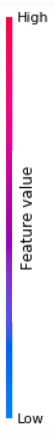}}
    \centering
    \begin{subfigure}[]{0.285\textwidth}
        \includegraphics[width=\textwidth]{Figures/crc_shap.png}
        \caption{Colorectal cancer}\label{fig:crc_shap}
    \end{subfigure}
    \qquad
    \begin{subfigure}[]{0.285\textwidth}
        \includegraphics[width=\textwidth]{Figures/liver_shap.png}
        \caption{Liver cancer}\label{fig:liver_shap}
    \end{subfigure}
    \qquad
    \begin{subfigure}[]{0.285\textwidth}
        \includegraphics[width=\textwidth]{Figures/lung_shap.png}
        \caption{Lung cancer}\label{fig:lung_shap}
    \end{subfigure}
    \begin{subfigure}[]{0.045\textwidth}
        \raisebox{6mm}{\includegraphics[width=\textwidth]{Figures/cohort_shap_legend.png}}
    \end{subfigure}
    \caption{Cohort-level SHAP summaries showing the contribution of the 15 laboratory markers with highest impact on the LR, separately for the (a) colorectal, (b) liver, and (c) lung cancer model. Red and blue points correspond to patients with high and low values of the corresponding laboratory markers, respectively. Laboratory markers are ordered along the y-axis with respect to their overall importance for the prediction, with most important markers at the top. ALP, alkaline phosphatase; AST, aspartate aminotransferase; BUN, blood urea nitrogen; MCH, mean corpuscular hemoglobin; MCV, mean corpuscular volume; RDW, red cell distribution width; WBC, white blood cells.}
    \label{fig:cohort_shap}
\end{figure}

To gain insights into the laboratory markers with highest impact on the LR, we show cohort-level SHAP summaries for our colorectal, liver, and lung cancer models in Figures \ref{fig:crc_shap}, \ref{fig:liver_shap}, and \ref{fig:lung_shap}, respectively. Since the LR can range from zero to infinity, we passed the computed LR for each sample through a logistic function with mean 5 and scale 0.5. This ensures that outliers, i.e., samples with significantly low or high LR, do not significantly impact the SHAP value distribution.

While only three (age, albumin and hematocrit) of the 15 most important laboratory markers are shared across all cancer types (only age among the five most important markers), there are at least seven markers shared between any two cancer types. % This supports our proposition to first pretrain a model on all cohorts to leverage as much data as possible that is relevant to all cancer types.
In contrast,  fine-tuned cancer type-specific prediction models are required to give more weight to markers predictive of only one cancer type, as, in fact, one (neutrophils), three (total protein, WBC and basophils (\%)), and three (BUN-creatinine ratio, calcium, and ALT) markers are most important in only colorectal, liver, and lung cancer, respectively.

Several of the biomarkers that are identified as important in our SHAP analysis have also been independently studied and reported. For instance, six of the 12 markers that were integrated into one or both sex-specific colon cancer prediction models by Goshen et al.\cite{Goshen2017} (ten of which are also available in our cohort) also appear to be most important in our CRC model (RDW, MCV, neutrophils, monocytes, hemoglobin, and AST).

For liver cancer, there are currently no screening recommendations, except for hepatitis B carriers, who are recommended by the American Association for the Study of Liver Diseases (AASLD) to start screening at age 40 and 50 years for men and women, respectively. In our analysis, platelet counts appear as the most important blood marker predictor of liver cancer. In fact, raised platelet counts (> 400 × $10^9$/l) indicative of thrombocytosis, or even high-normal platelet counts (326–400 × $10^9$/l) have been reported to be associated with higher cancer incidence, in particular colorectal and lung cancer\cite{Mounce2020}. However, patients with high-normal platelet counts had, in general, advanced-stage cancer at diagnosis, which may explain why platelets are particularly important in our liver cancer model, as liver cancer is mostly detected at a late stage, emphasizing the need for improved guidelines concerning screening eligibility.

Of note, the lack of a clear color gradation indicates interactions between different laboratory markers, illustrating the need to train a holistic model by integrating multiple laboratory markers.

\subsection*{Analysis by comorbidities}
\begin{figure}
    \centering
    \begin{subfigure}[]{\linewidth}
        \includegraphics[width=\linewidth]{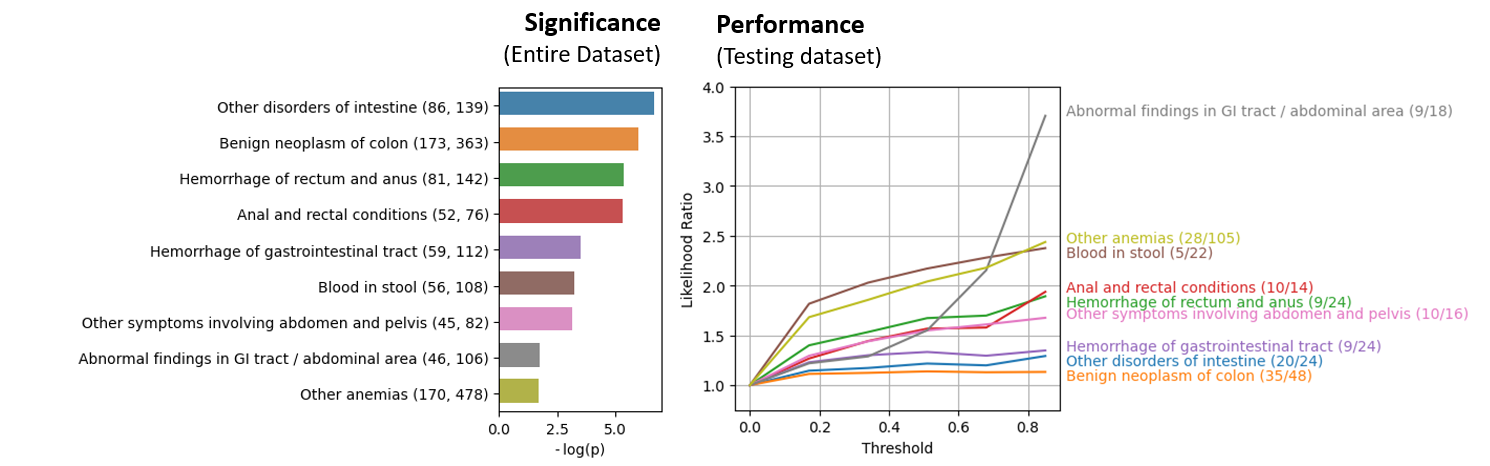}
        \caption{Colorectal cancer}\label{fig:crc_comorb}
    \end{subfigure}
    \begin{subfigure}[]{\linewidth}
        \includegraphics[width=\linewidth]{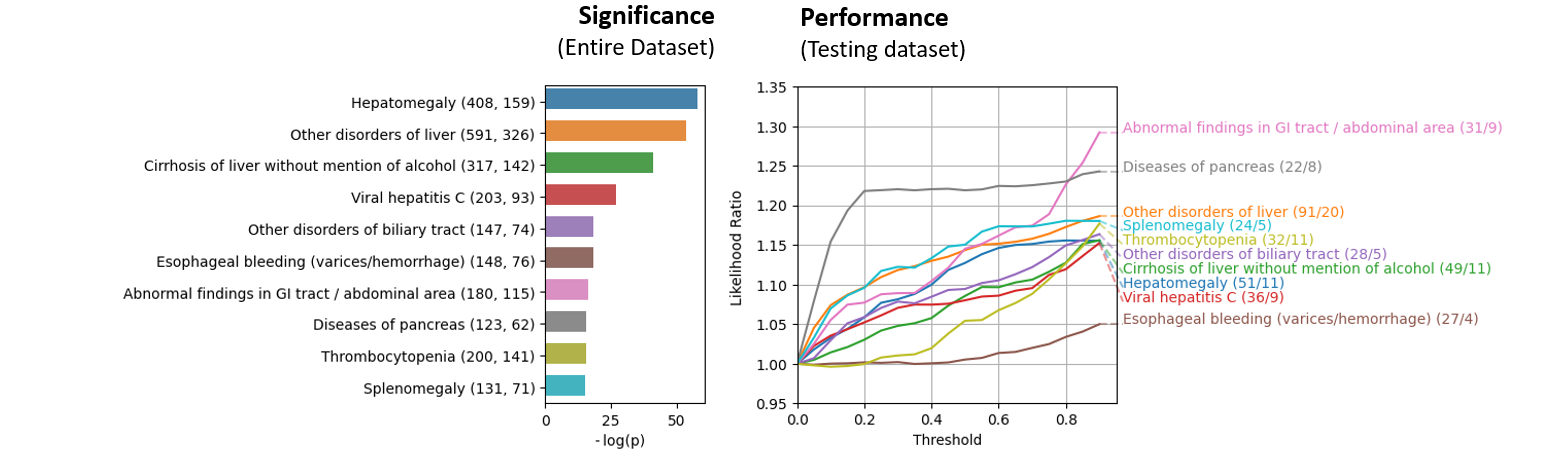}
        \caption{Liver cancer}\label{fig:liver_comorb}
    \end{subfigure}
    \begin{subfigure}[]{\linewidth}
        \includegraphics[width=\linewidth]{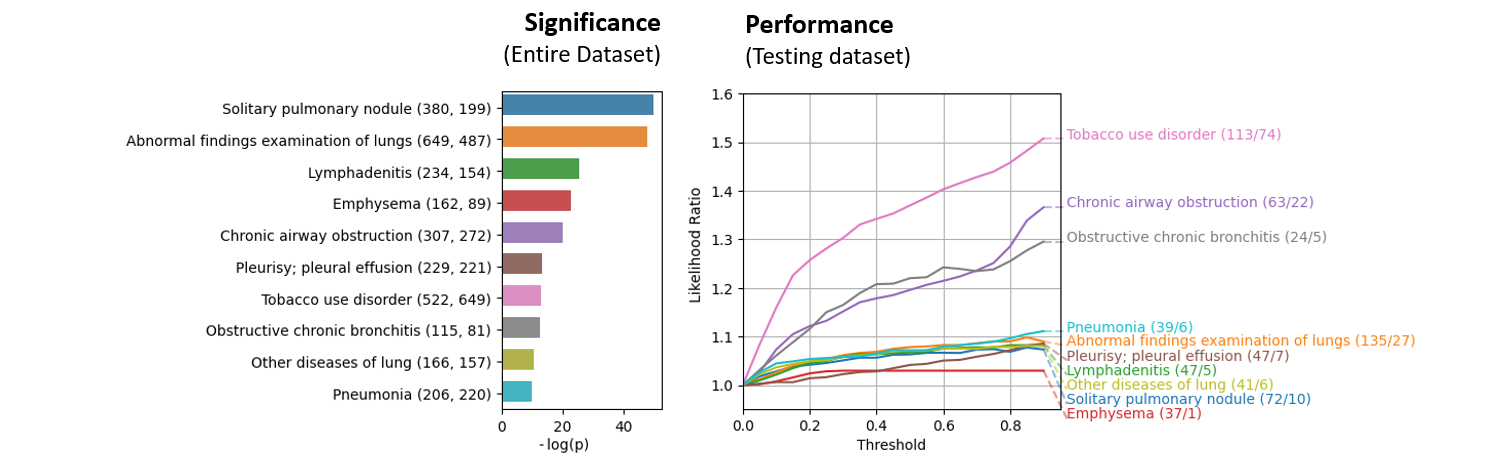}
        \caption{Lung cancer}\label{fig:lung_comorb}
    \end{subfigure}
    \caption{Comorbidity analysis on (a) colorectal, (b) liver, and (c) lung cancer validation cohorts. Likelihood ratio plots show the likelihood ratios of corresponding models on subgroups of patients suffering from the indicated comorbidities, for increasing risk thresholds on the testing dataset. Barplots illustrate the significance of association of the corresponding comorbidity with cancer type (as negative logarithm of the p-value obtained with Fisher's exact test) over the entire dataset. The prevalence of the respective comorbidities in the cancer-positive and control cases is listed with each comorbidity. Note that the base prevalence is different for each comorbidity. GI, gastrointestinal.}
    \label{fig:comorb}
\end{figure}

For each of the cancer cohorts, we used phecodes\cite{Wu2019} to identify comorbidities that are more likely to be present in the underlying population prior to cancer diagnosis. Relative odds ratios were computed to select comorbidities that are significantly different between cancer and control samples using Fisher's exact test (Figures \ref{fig:crc_comorb}, \ref{fig:liver_comorb}, and \ref{fig:lung_comorb}). We found that those are usually the conditions or symptoms that raise suspicion of cancer or other organ dysfunction in a cancer population. For instance, the top three significant comorbidities prior to diagnosis in the CRC cohort are other disorders of the intestine, benign neoplasm of the colon, and hemorrhage of rectum and anus (Figure \ref{fig:crc_comorb}).

Since the FIT test screens stool for occult blood from the lower intestines, we use the subgroup of patients with blood in stool as a surrogate to evaluate the added value of our model in comparison to an established screening test. Compared to the base prevalence, we see an increase in LR greater than 3-fold, at a risk threshold greater than 0.8 (Figure \ref{fig:crc_comorb}). Furthermore, we also see added value in subgroups of patients with conditions such as hemorrhage of rectum and anus or gastrointestinal tract.

For lung cancer, the USPSTF criteria for screening comprise age and smoking history\cite{USPSTF2021_Lung}. While our lung cohort does not include smoking history, we analyzed the performance of our model on a subgroup of patients having a tobacco use disorder. Notably, our model shows added value in identifying high-risk patients (more than 50 \% increase in LR as compared to the base prevalence, at a risk threshold greater than 0.9, Figure \ref{fig:lung_comorb}).

% \subsection*{Subgroup analysis w.r.t. lipid panel}
% Analysis on patient subgroups stratified by lipid panel measurements such as low vs. high HDL cholesterol.

% \subsubsection*{Uncertainty estimation methods}
% MC Dropout
% Will a shared model help with the rejection of outlier cases
% Ensemble vs Gaussian process models?

\section*{Discussion}
Here, we report on the development and evaluation of separate deep learning-based risk prediction models for three cancer types based on routine laboratory marker measurements. Although other blood marker-based pan-cancer or cancer-specific risk prediction models have been reported, a direct comparison of performance and markers identified to be important for risk assessment is difficult due to varying (i) study designs, e.g., with respect to the follow-up period until cancer diagnosis (90 days\cite{Soerensen2022} vs. 365 days in our model), (ii) cancer prevalence in training and validation cohorts (25.2\% in\cite{Flyckt2024} vs. 5.7 \% in our lung cohort), (iii) inclusion criteria (e.g., smoking history in lung cancer), (iv) number and availability of markers (e.g., restrict to samples with complete data vs. imputation of missing data), (v) inclusion of non-routine blood markers (e.g., AFP and CA-125\cite{Naeser2017}) as well as patient characteristics other than age and sex\cite{Gould2021}. While the performance is not yet adequate enough for the model to be used as a diagnostic test, we anticipate it to be a valuable tool for recruiting patients into screening programs and, thus, increase screening uptake and efficiency overall, pending additional confirmatory studies. In future work, we also plan to adopt the model to personalize the screening interval after a negative screening result (e.g., based on LDCT).

While the current USPSTF guideline for lung cancer screening only considers age and smoking history for screening eligibility\cite{Jonas2021}, a multitude of risk prediction models have been proposed and reported to show improved performance over the USPSTF criteria\cite{Toumazis2020}. For instance, the PLCO\textsubscript{m2012} model includes race, body mass index, education, presence of chronic lung disease, personal history of cancer, and family history of lung cancer, in addition to age and smoking history\cite{Tammemagi2013}. Furthermore, a four-marker protein panel (4MP), measuring a precursor form of surfactant protein B (Pro-SFTPB), cancer antigen 125 (CA-125), carcinoembryonic antigen (CEA), and cytokeratin-19 fragment (CYFRA 21-1), has been combined with PLCO\textsubscript{m2012} to predict lung cancer risk\cite{Fahrmann2022}. However, the required biomarkers are not assessed within routine blood testing, posing a challenge for the test's clinical utility. In contrast, our model is only based on routine laboratory markers and even performs comparable to models that integrate knowledge about smoking status or smoking history\cite{Flyckt2024, Gould2021}.

As cohorts were enriched with patients diagnosed with chronic liver disease, the better performance of our model in liver cancer might originate from an improved differential diagnosis. Interestingly, ALT and AST are not picked up as important markers for liver cancer either, which might also be tied to the model trying to differentiate from chronic liver disease cases in the control set. Analogous, the addition of cases with chronic bowel and lung disease could prove helpful in the colorectal and lung cancer setting, respectively.

The SHAP analysis in Figure \ref{fig:cohort_shap} depicts the contributions of important laboratory markers to the normalized LR on the cohort level. In addition, we provide the relative contributions of laboratory markers to model prediction on the individual level as waterfall plots in Supplementary Figure S2 for two selected cases: (1) a liver cancer case with only four OoR markers (total CO\textsubscript{2}, lymphocytes (\%), hemoglobin, and hematocrit, Supplementary Figure S2a), of which total CO\textsubscript{2} and hematocrit are the only 2 of the 15 laboratory markers with highest impact on model prediction (Figure \ref{fig:liver_shap}), and a control case with 18 OoR markers (Supplementary Figure S2b), both correctly classified by the model with a high ($>0.8$) and low ($\sim0$) risk, respectively. These example cases illustrate the importance of considering the measurements of a range of markers together to achieve optimal risk assessment.

In this work, we have focused on CBC and CMP blood tests as availability of complete lipid panel blood tests was limited in our cohorts. While lipid panel use is less common in Europe, it is more frequently ordered in the US. In fact, we expect its integration into our risk models to further improve cancer risk assessment by contributing complementary information on physiological or pathological status.

To avoid potential harm resulting from unnecessary follow-up procedures, differential life expectancy, e.g., as a result of competing comorbidities, needs to be taken into account\cite{Wu2023}. This is of particular importance, as chronic diseases may increase the risk of complications from biopsies and cancer treatment. While we have focused on prioritizing patients for screening eligibility, allowing clinicians to discuss with patients who have been predicted to be at increased risk to conduct follow-up screening procedures as early as possible, we plan to investigate different risk grading and threshold approaches that may be applied in dependence of the presence/absence of certain comorbidities.

Similarly, it would be of great value if we could explore the contributions of laboratory markers with respect to cancer stage. Unfortunately, we did not have access to a cohort-linked cancer registry and, therefore, lack information on cancer stage. In the future, we anticipate such an analysis to open up the possibility to develop cancer risk prediction models that also provide an assessment of cancer progress.

%Label inconsistencies based on the data extracted from medical records are baked into the model. Since large-scale one-by-one consistency checks and chart reviews are laborious, we still need technological approaches to be developed to address this challenge.

% TODO: Talk about coding inaccuracies as limitation of the study, hence only perform analysis where coded. However, chronic conditions may have false alarms, so additional validation with curated datasets is additionally required.
% https://www.ncbi.nlm.nih.gov/pmc/articles/PMC7939013/
% https://academic.oup.com/jpubhealth/article/34/1/138/1553464
% https://bmcmedinformdecismak.biomedcentral.com/articles/10.1186/s12911-021-01531-9

\section*{Methods}
\subsection*{Study selection}
We created a cohort of individuals diagnosed with malignant neoplasms of colorectal, liver or lung, or had no confirmed diagnosis of any malignant neoplasm, within the period from 2017 to 2021. We further enriched the cohort by considering additional patients who are diagnosed with chronic kidney disease and/or chronic liver disease but have no personal history or a novel cancer diagnosis during the aforementioned period. The cohort was created using the Prognos Factor® platform, and the corresponding patient records were obtained from Prognos Health. The patient records include the laboratory marker measurements (blood or urine based) from various encounters as well as CPT\cite{CPT2023} and ICD-10 codes\cite{ICD-10-CM2022} from claims. ICD-10 codes of patients were used as surrogate for the diagnosis for selecting patients in this study. All the records obtained correspond to an anonymized healthcare provider in the US.
% Given the longitudinal record of each patient, the cancer diagnosis was determined from the ICD-10 codes recorded for each patient. If a patient has an ICD-10 code of C18, C19, C20, indicating neoplasm of either colon or rectum then a diagnosis label of Colorectal Cancer is assigned\cite{ICD-10-CM2022}. Similarly, diagnosis labels of Liver and Lung cancer are assigned to patients with ICD-10 codes of C22 and C34, respectively. Control patients are the patients who do not have any malignant neoplasm within the 12-month based on ICD-10 code (i.e., exclude patients with ICD-10 codes starting with C). 

\subsection*{Screening-based cohorts}
To create cohorts for the pre-screening scenario, we utilized the medical claims data to select the patients and their visits that correspond to a screening procedure. Table \ref{tab:cohort_selection} lists the CPT and ICD-10 codes used for each cancer type. If screening-related codes were not reported for a patient, we used diagnostic procedure codes (listed in Table \ref{tab:cohort_selection} as well).

%\begin{table}[ht]
%\centering
%\begin{tabular}{|l|c|c|c|c|}
%\hline
%Cancer type & \makecell[l]{Screening procedure \\codes} & \makecell[l]{Screening encounter \\ codes}
%& \makecell[l]{Diagnostic procedure \\ codes} & \makecell[l]{Diagnosis codes} \\
%\hline
%\hline
%Colorectal cancer &
%\makecell[l]{G0105, G0120, G0121, \\ G2204, 45378, 45388, \\ 45330, 45381} &
%\makecell[l]{Z1211, Z1212, \\ Z1213} &
%\makecell[l]{45380, 45382, 45384, \\ 45385, 45390} &
%C18, C19, C20  \\
%\hline
%\makecell[l]{Hepatocellular \\ carcinoma} &
%76700, 76705, 78215 &
%Z1289 &
%\makecell[l]{47000, 74176, 74177, \\ 74148} &
%C22 \\
%\hline
%Lung cancer &
%\makecell[l]{G0296, 71271, \\X-ray (71045, 71046, \\ 71047, 71048)*} &
%Z122 &
%71250 &
%C34 \\
%\hline
%\end{tabular}
%\caption{\label{tab:cohort_selection} Screening and diagnosis relevant ICD-10 and CPT codes used %for patient selection and assigning diagnosis labels for each cancer.}
%\end{table}

\begin{table}[ht]
\renewcommand*{\arraystretch}{1.4}
\begin{tabular}{C{0.11\linewidth} C{0.23\linewidth} C{0.15\linewidth} C{0.23\linewidth} C{0.15\linewidth}}
\toprule
\multirow{2}{*}{\textbf{Cancer type}} & \multicolumn{2}{c}{\textbf{Screening}} & \multicolumn{2}{c}{\textbf{Diagnosis}} \\
\cmidrule(lr){2-3}
\cmidrule(lr){4-5}
& \textbf{Procedure codes} & \textbf{Encounter codes} & \textbf{Procedure codes} & \textbf{ICD-10 codes} \\
\midrule
Colorectal & \makecell{G0105, G0120,\\ G0121, G2204, 45378,\\ 45388, 45330, 45381} & \makecell{Z1211,\\ Z1212, Z1213} & \makecell{45380, 45382,\\ 45384, 45385, 45390} & C18, C19, C20 \\
Liver & 76700, 76705, 78215 & Z1289 & \makecell{47000, 74176,\\ 74177, 74148} & C22 \\
Lung & \makecell{G0296, 71271,\\ X-ray (71045, 71046, \\ 71047, 71048)\footnotesize$^*$} & Z122 & 71250 & C34 \\
\midrule
\multicolumn{5}{l}{\footnotesize$^*$ Reimbursement as a screening procedure was discontinued in late 2018.} \\
\bottomrule
\end{tabular}
\caption{Screening and diagnosis relevant CPT and ICD-10 codes used for patient selection and assignment of diagnosis labels for each cancer.}
\label{tab:cohort_selection}
\end{table}

For each patient in the screening cohorts, we next determined if the patient had a positive cancer diagnosis. Although we used ICD-10 codes to select patients for the study from the Prognos Factor® platform, we recognized that the use of ICD-10 codes as confirmed diagnosis may not be sufficiently reliable\cite{Campbell2019}. Therefore, we ensured that at least one additional CPT or ICD-10 code associated with cancer diagnosis and/or therapy (chemotherapy and/or radiation therapy) is present after diagnosis. Figures \ref{fig:consort_crc}, \ref{fig:consort_liver}, and \ref{fig:consort_lung} show the respective consort flow diagrams for each cancer. The screening cohorts for each cancer type as described above correspond to the first cell in the flow diagram for each cancer.

\begin{figure}
\centering
\begin{subfigure}[t]{\linewidth}
\includegraphics[width=\linewidth]{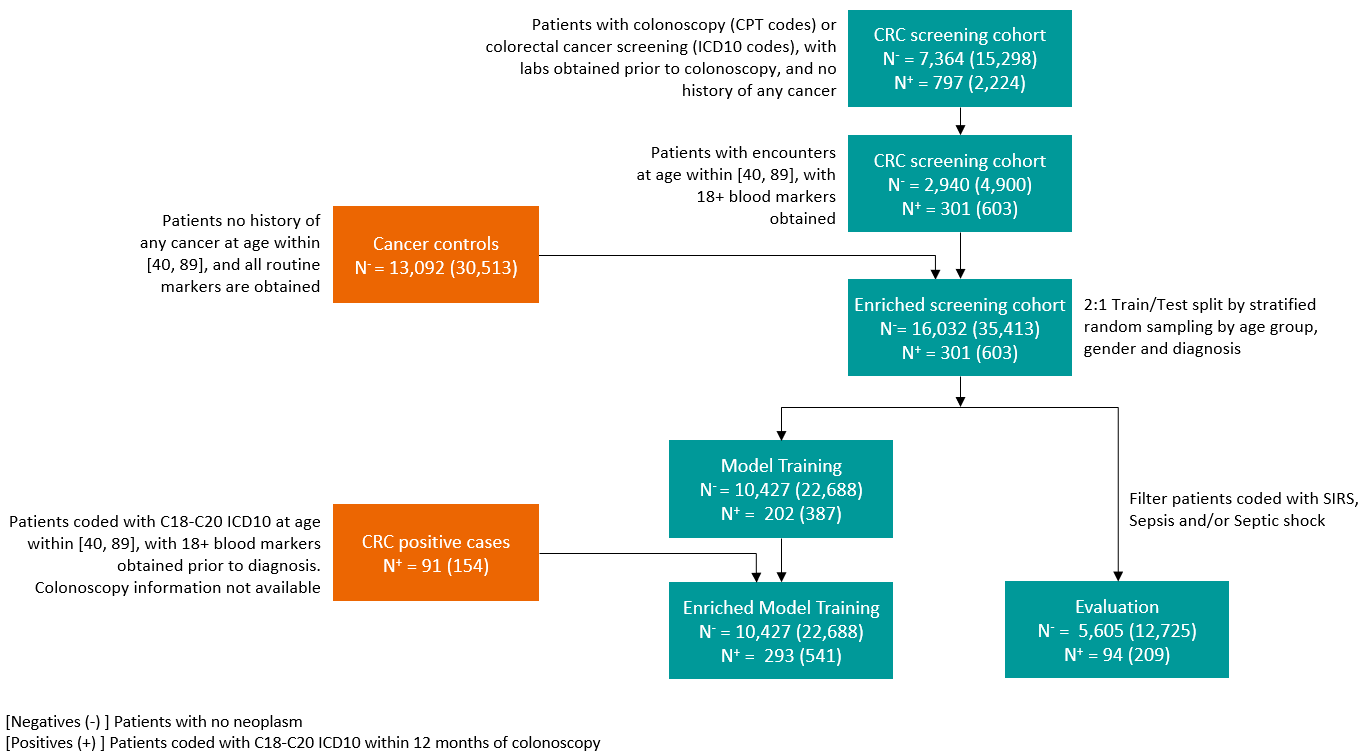}
\caption{Colorectal cancer}\label{fig:consort_crc}
\end{subfigure}

\begin{subfigure}[b]{\linewidth}
\includegraphics[width=\linewidth]{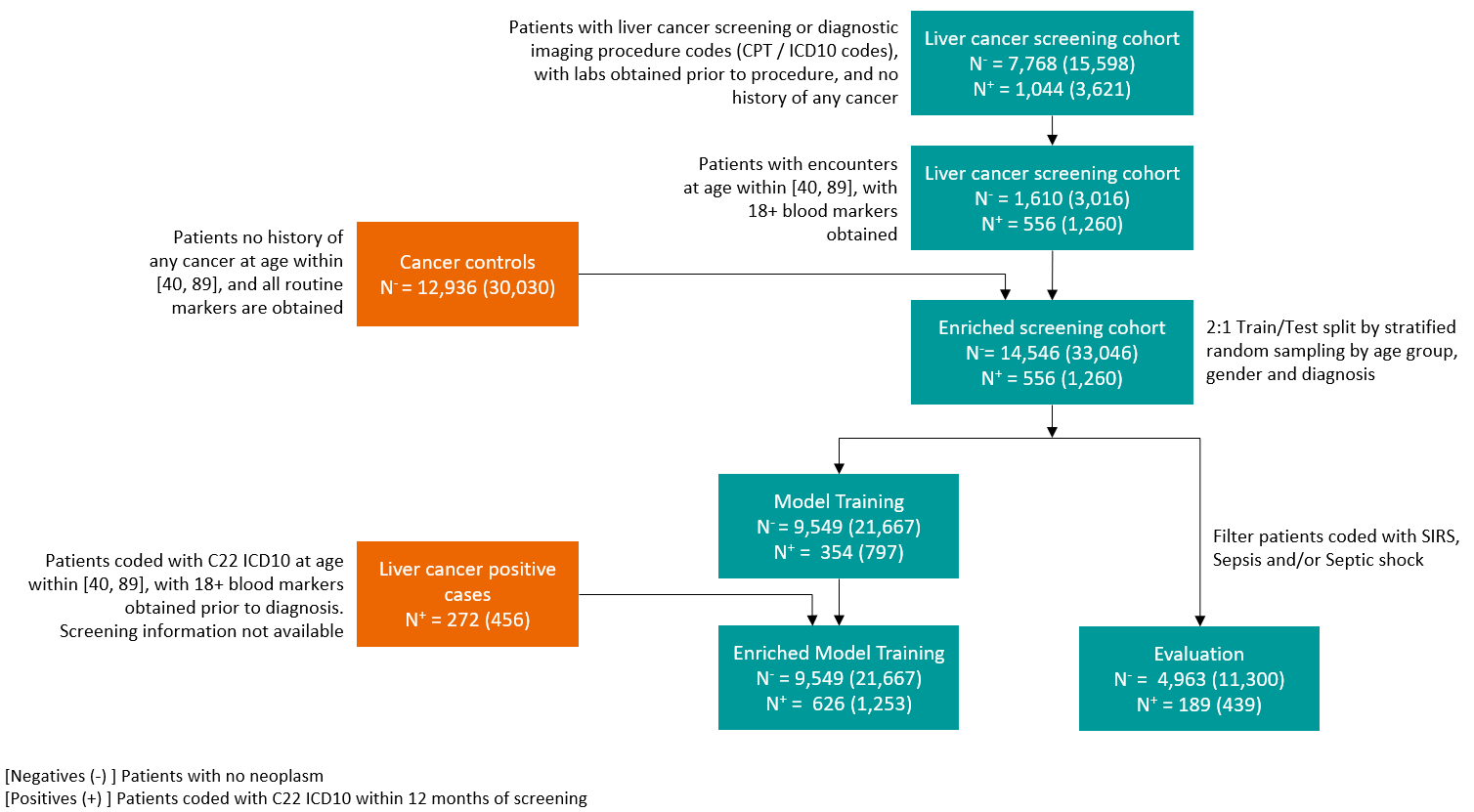}
\caption{Liver cancer}\label{fig:consort_liver}
\end{subfigure}
\end{figure}

\begin{figure}\ContinuedFloat
\begin{subfigure}[b]{\linewidth}
\includegraphics[width=\linewidth]{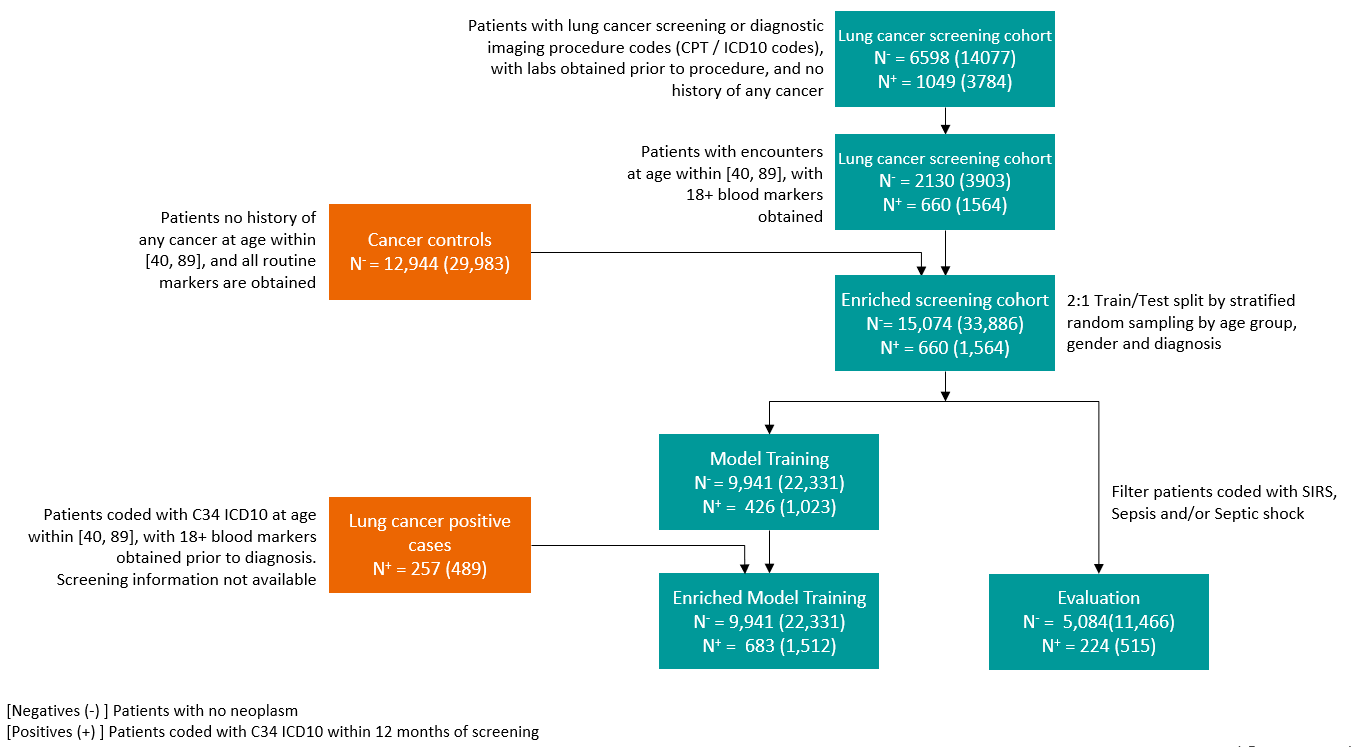}
\caption{Lung cancer}\label{fig:consort_lung}
\end{subfigure}
\caption{Consort flow diagrams for various cancer cohorts. Each cell depicts the number of cancer positive (N+) and cancer negative (N-) patients. For each entry, both the number of patients and their aggregate encounter count has been reported. SIRS, systemic inflammatory response syndrome.}
\label{fig:consort}
\end{figure}

For patients with a positive cancer diagnosis, we only considered the records from visits within 12 months prior to the visit associated with the screening/diagnostic procedure. This ensured that the records all had a cancer diagnosis within the next 12 months. For patients with no cancer, we only considered records for which there are subsequent records available. For each cell in the consort flow diagrams, both the aggregate patient and record counts are shown.

To be consistent with the pre-screening scenario, we next identified the records of all selected patients between the age of 40 and 89, with no prior diagnosis of cancer. The laboratory markers obtained at each visit are tied to the underlying conditions that were diagnosed or monitored and, hence, the number and type of markers varied significantly, e.g., from a single marker to all the markers in CBC and CMP. Thus, we only considered visits with at least 18 markers for our analysis and discarded all other visits. While this approach potentially results in some data loss, it helps avoid the machine learning model to overfit to the specific markers measured. Note that CBC and CMP also include derived markers such as BUN-creatinine ratio which, if missing in the records, are computed using the reported BUN and creatinine values. Similarly, blood-count markers and their percentage markers were computed if either one was not reported.

We subsequently enriched the resulting cohorts by including records of additional patients between the age of 40 and 89 who have no history of cancer and no cancer diagnosis within the next 12 months. While this step enriched the data for model development, more importantly, it also incorporated individuals who are eligible for screening but may not have gone through screening within the period considered in this study. We note that the number of patients (records) that were used to enrich each cancer cohort is not the same in the consort flow diagram. This is primarily due to the fact that there are patients that have gone through a screening procedure for one or more cancers but not all.

\subsection*{Data preparation and pre-processing}
Given all the patients and their records in the enriched screening cohorts, we next split the data into development and validation cohorts in a 2:1 ratio, stratified by age, sex, and diagnosis. We then performed additional steps to enrich the development cohort that may assist in training a robust model as well as in cleaning the validation cohort to remove any ambiguous records. Specifically, we filter out patient records that correspond to severe/acute infection such as systemic inflammatory response syndrome (SIRS), sepsis, and/or septic shock, since they can result in significant change in CBC and CMP marker levels and may not be relevant in the pre-screening scenario. We also enriched the development cohort by adding newly diagnosed cancer patients whose records do not include a visit associated with any of the screening codes.

Given the development and validation cohorts, we normalized the distributions of biomarker values by subtracting their median and dividing by their inter-quartile distance (IQD). In addition, we transformed the distributions of the following markers to the logarithmic (log\textsubscript{10}) scale prior to their normalization: alanine aminotransferase (AST), alkaline phosphatase (ALP), aspartate aminotransferase (AST), bilirubin, creatinine, erythrocyte distribution width (RDW), glucose, leukocytes, lymphocytes, neutrophils, and urea nitrogen. Note that the pre-processing parameters (i.e., median and IQD values) for all markers were only computed on the development cohort, and the same parameter values were then applied to the validation cohort. 

\subsection*{Deep Profiler}
Given the pre-processed development cohort, we employed a deep neural network called Deep Profiler\cite{Singh2021} to train models for each of the three cancer scenarios. This method has previously been applied to train models to predict a severity progression risk score for COVID-19 patients based on biomarkers (age and nine blood markers) obtained at the time of admission\cite{Singh2021, Yilmaz2024}. Furthermore, it was demonstrated to be robust to marker missingness as well as competitive to other methods such as a logistic regression and boosted forests.

The Deep Profiler model builds on a variational autoencoder architecture\cite{Pinheiro2021}, and consists of three main networks: an encoder for extracting prominent features represented in a latent space, a decoder for reconstructing the input data to ensure data fidelity of the latent feature representation, and, finally, a multi-label classifier network, which is trained to estimate the ordinal risk score. Use of autoencoders to deal with robustness to missing data has been demonstrated to be effective\cite{Smieja2019, Pereira2020}.

In this work, we use the same Deep Profiler architecture for each of the three cancer models, with a symmetric encoder-decoder structure, each comprising of blocks of three fully connected layers with 32 kernels. Each fully connected layer in the encoder is followed by a batch normalization and LeakyReLU (0.2), and each fully connected layer in the decoder is followed by batch normalization and ReLU. Network parameters were learned using the ADAM optimizer with an initial learning rate of 10\textsuperscript{-4}. The training loss is a combination of reconstruction loss (only applied to corresponding input features whose values are not missing), VAE regularization loss and binary cross entropy loss based on the patient's diagnosis.

\subsection*{Prediction uncertainty using Deep Profiler ensembles}
While the VAE helps adding robustness to data missingness, undesirable biases may still be present due to the patient population/sampling. To this end, we trained an ensemble of Deep Profiler models. Given the ensemble of models, a confidence interval for the risk score (output of the sigmoid layer) is computed. Each model is trained on different subsets of the data, accounting for variations in the acquisition protocols across hospitals, availability of various biomarkers, measurement methods used to assess the biomarker as well as biomarker measurements such as laboratory values, age range, comorbidities, etc. In this study, for each cancer scenario, we used an ensemble of ten Deep Profiler models.

\subsection*{Estimation of likelihood ratios}
Given the patient biomarker data and computed confidence interval over the risk score, the next key step is to present this information such that it can be easily interpreted and acted upon. To this end, we used the risk score and confidence interval to identify the subgroup of patients in the development cohort with similar scores. As a result, we obtained a “similar patient cohort” (a.k.a. patient cohort with similar risk scores) in addition to the development cohort. We then computed the pre- and post-test odds for the patient to have the disease and, subsequently, calculated the likelihood ratio. This can be done separately for each medical condition. Figure \ref{fig:inference} below shows one way of presenting this information to the clinician.

\begin{figure}[ht]
\centering
\includegraphics[width=\linewidth]{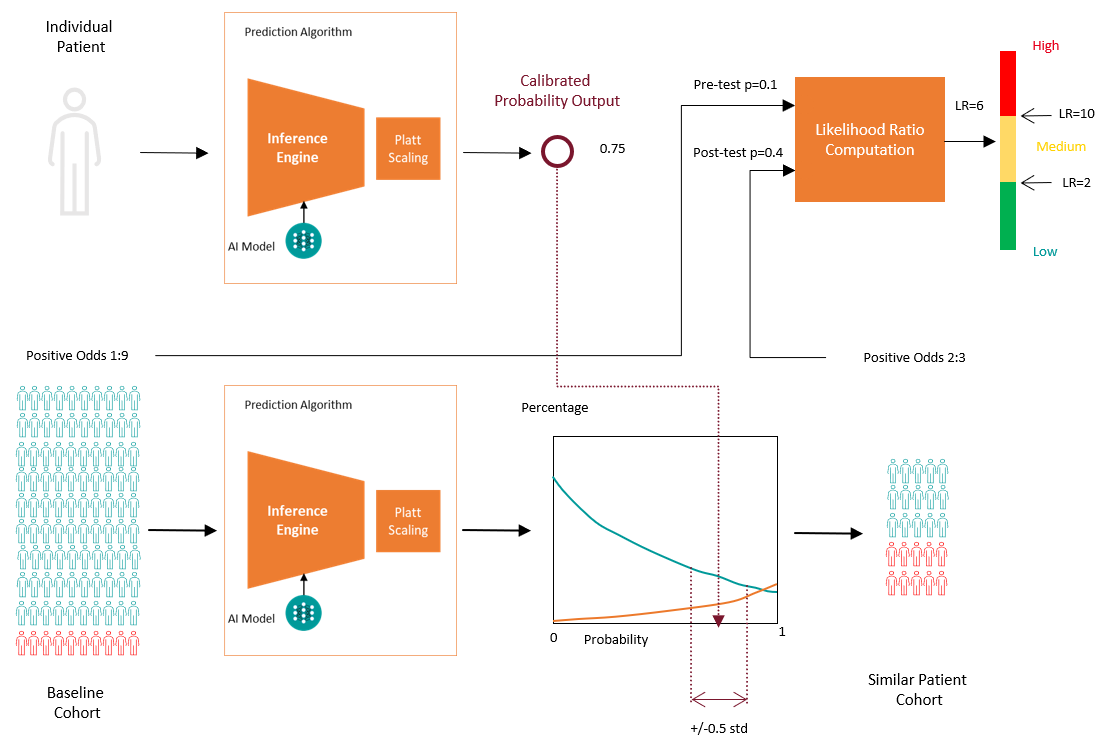}
\caption{Schematic of the system output interpretation using likelihood ratio.}
\label{fig:inference}
\end{figure}

% based on the pre-test probability based on the disease prevalence, and post-test probability based on the disease prevalence in the 

% \subsection*{Identifying relevant features for risk assessment}
% Describe the shapley method performed over 

\subsection*{Subgroup analysis based on comorbidities }
We used phecodes\cite{Wu2019} to analyze comorbidities in each of the cohorts. Phecodes are manually curated groups of ICD-10 codes that are clinically meaningful and relevant to research. They have been widely applied in phenome-wide association studies but, more recently, were also used for rapid electronic health record phenotyping\cite{Bastarache2021, Kerley2022}. For the colorectal, liver and lung cancer cohorts, we censored all ICD-10 codes assigned after the date of diagnosis and only mapped the codes assigned before diagnosis to phecodes. This ensured the identification of clinical comorbidities prior to a cancer diagnosis. For control cohorts, we mapped all of the ICD-10 codes recorded to phecodes. Subsequently, we compared the counts of individuals with each phecode in the control group against each of the pre-diagnosis cancer cohorts and identified significantly different phecodes, or comorbidities, based on Fisher's exact tests. Finally, we ranked the most significant comorbidities with at least 50 individuals in both the cancer and the control cohort ranked for subgroup analysis and evaluated the respective Deep Profiler models in those subgroups.

\section*{Data availability}
Entire patient cohort data used in this study was licensed from Prognos Health (prognoshealth.com) via the prognosFACTOR® platform. Prognos Health can be reached at https://prognoshealth.com/support or via email at client\_support@prognoshealth.com. Kindly reach out to the corresponding author at vivek-singh@siemens-healthineers.com if you are interested in additional details of the patient selection criteria used to obtain the records.

\section*{Code availability}
The inference algorithms of the models presented in this paper are available from the authors upon request.

\bibliography{oncoscreen}
% \noindent LaTeX formats citations and references automatically using the bibliography records in your .bib file, which you can edit via the project menu. Use the cite command for an inline citation, e.g. \cite{Hao:gidmaps:2014}.
% For data citations of datasets uploaded to e.g. \emph{figshare}, please use the \verb|howpublished| option in the bib entry to specify the platform and the link, as in the \verb|Hao:gidmaps:2014| example in the sample bibliography file.

\section*{Acknowledgements}
We would like to thank Dennis Gilbert, Rangarajan Sampath, and Harigovind Singh for valuable discussions about this work. We would like to also thank Manish Chowdhury for assisting us with data preparation during the initial stage of development.

\section*{Author contributions statement}
V.S., S.C., M.S., R.G., J.G., D.C., and A.K. conceived the model and experiments, V.S. implemented the model, V.S., S.C., M.S., S.R. and A.P. prepared the data, V.S., S.C., and M.S. conducted the experiments, V.S., S.C., M.S., and A.K. analyzed the results, V.S., S.C., M.S., and A.K. wrote the manuscript, which was then reviewed and edited by all authors.

\section*{Additional information}

\noindent\textbf{Competing interests}

\noindent V.S., S.C., M.S., J.G., D.C., and A.K. are employed by Siemens Healthineers. A.P. is employed by Siemens SRL. S.R. and R.G. were employed by Siemens Healthineers during the completion of this work.

\vspace{\baselineskip}
\noindent\textbf{Disclaimer}

\noindent The concepts and information presented in this paper are based on research results that are not commercially available. Future commercial availability cannot be guaranteed.

\end{document}

\flushbottom
\maketitle

\thispagestyle{empty}

\section*{Supplementary Figures}

\renewcommand{\thefigure}{S\arabic{figure}}
\renewcommand{\figurename}{Supplementary Figure}
\setcounter{figure}{0}

\renewcommand{\thetable}{S\arabic{table}}
\renewcommand{\tablename}{Supplementary Table}
\setcounter{table}{0}

\begin{figure}[ht]
    \centering
    \begin{subfigure}[]{0.3\textwidth}
        \includegraphics[width=\textwidth]{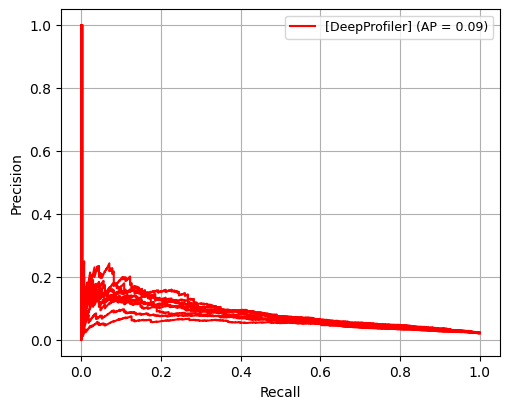}
        \caption{Colorectal cancer}\label{fig:crc_pr}
    \end{subfigure}
    \qquad
    \begin{subfigure}[]{0.3\textwidth}
        \includegraphics[width=\textwidth]{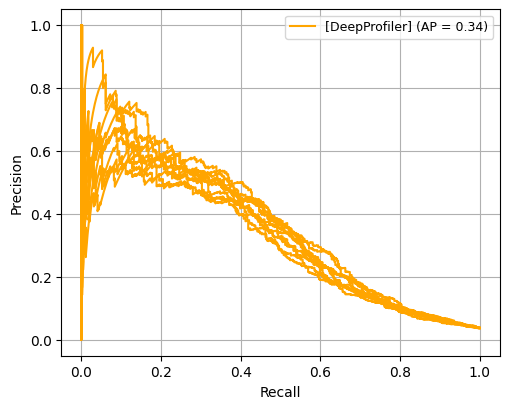}
        \caption{Liver cancer}\label{fig:liver_pr}
    \end{subfigure}
    \qquad
    \begin{subfigure}[]{0.3\textwidth}
        \includegraphics[width=\textwidth]{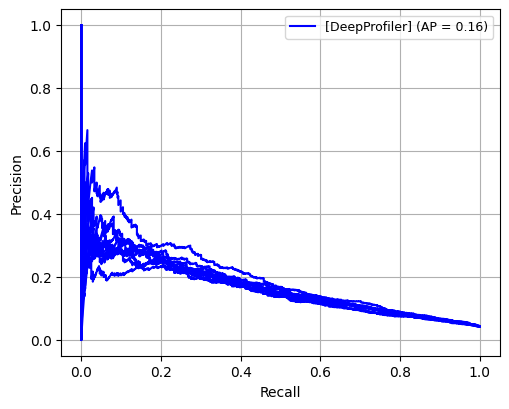}
        \caption{Lung cancer}\label{fig:lung_pr}
    \end{subfigure}
    \caption{Quantitative performance assessment on (a) colorectal, (b) liver, and (c) lung cancer validation cohorts. Precision-recall curves represent the performance of single Deep Profiler models constituting the respective ensemble. The average precision (AP) corresponds to the weighted mean of precisions achieved at each threshold, with the increase in recall from the previous threshold used as the weight.}
    \label{fig:performance_pr}
\end{figure}

\begin{figure}[ht]
    \centering
    \begin{subfigure}[]{0.46\textwidth}
        \includegraphics[width=\textwidth]{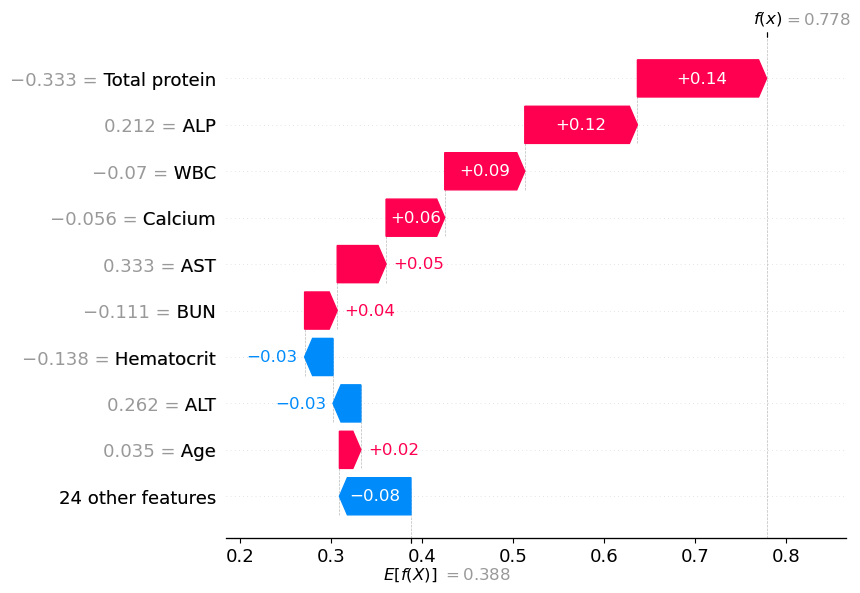}
        \caption{Liver cancer case}\label{fig:liver_positive_shap}
    \end{subfigure}
    \qquad
    \begin{subfigure}[]{0.46\textwidth}
        \includegraphics[width=\textwidth]{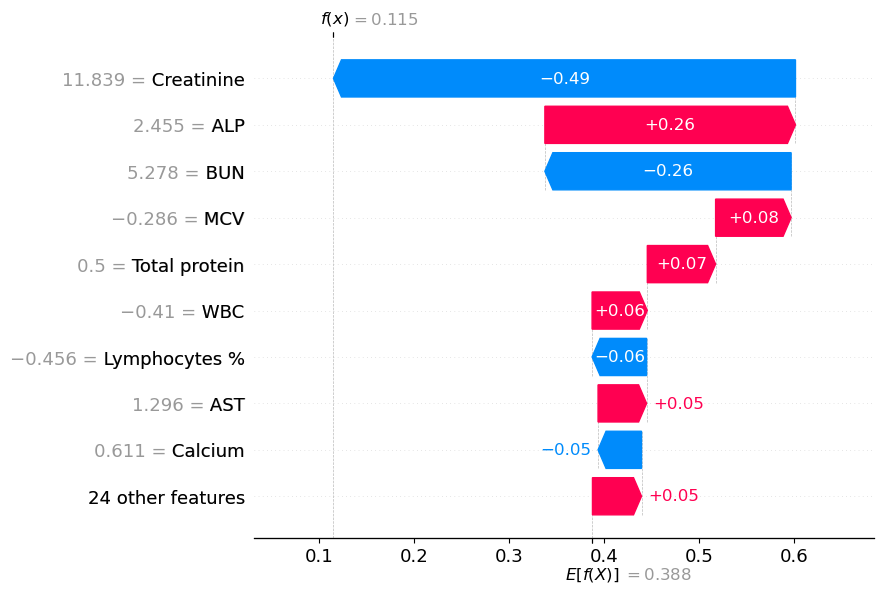}
        \caption{Control case}\label{fig:liver_negative_shap}
    \end{subfigure}
    \caption{Waterfall plots of SHAP feature attributions showing the contribution of the nine laboratory markers with highest impact on the normalized LR (f(x)) for (a) a liver cancer and (b) a control case. Plots are aligned at the average normalized LR (E[(f(x)]) across all samples. Samples had to have data available for at least 24 laboratory markers. Normalized laboratory values are listed along with the marker names.}
    \label{fig:liver_shap_waterfall}
\end{figure}

\begin{table*}[ht]
\renewcommand*{\arraystretch}{1.4}
\begin{center}
\begin{small}
\begin{tabular}{rcccc}
\toprule
\textbf{Characteristics} & \textbf{No cancer} & \textbf{Colorectal cancer} & \textbf{Liver cancer} & \textbf{Lung cancer} \\ 
\midrule
Encounters & 34062 & 750 & 1559 & 1702 \\ 
Age & 61.38 $\pm$ 12.28 (17390) & 63.11 $\pm$ 10.82 (387)* & 64.05 $\pm$ 11.21 (751)* & 67.48 $\pm$ 10.03 (752)* \\ 
Sex: Male & 8473 (17390) & 196 (387) & 434 (751)* & 384 (752) \\ 
\midrule
Albumin (g/dL) & 4.5 $\pm$ 0.3 (16329) & 4.2 $\pm$ 0.4 (239)** & 4.0 $\pm$ 0.6 (631)** & 4.2 $\pm$ 0.4 (704)** \\ 
ALP (U/L) & 80.8 $\pm$ 45.0 (16790) & 96.5 $\pm$ 52.4 (311)** & 166 $\pm$ 169 (704)** & 103 $\pm$ 77 (713)** \\ 
ALT (U/L) & 24.6 $\pm$ 24.2 (16273) & 28.0 $\pm$ 24.8 (248) & 44.8 $\pm$ 74.9 (611)** & 21.6 $\pm$ 18.1 (708)** \\ 
AST (U/L) & 24.0 $\pm$ 23.4 (16319) & 29.4 $\pm$ 20.6 (254)** & 46.7 $\pm$ 54.6 (620)** & 24.2 $\pm$ 18.6 (698) \\ 
Bilirubin (mg/dL) & 0.5 $\pm$ 0.4 (16745) & 0.5 $\pm$ 0.3 (313)* & 1.0 $\pm$ 2.0 (684)** & 0.4 $\pm$ 0.4 (722)** \\ 
BUN (mg/dL) & 16.4 $\pm$ 8.3 (16681) & 16.2 $\pm$ 12.0 (284)** & 15.9 $\pm$ 7.3 (642)* & 17.7 $\pm$ 9.9 (721)** \\ 
BUN-Creatinine Ratio & 17.6 $\pm$ 5.6 (14545) & 17.3 $\pm$ 6.0 (264) & 18.1 $\pm$ 6.0 (599)* & 18.0 $\pm$ 6.3 (716) \\ 
Calcium (mg/dL) & 9.4 $\pm$ 0.4 (16873) & 9.3 $\pm$ 0.5 (310)** & 9.2 $\pm$ 0.6 (689)** & 9.3 $\pm$ 0.6 (723)** \\ 
Chloride (mEq/L) & 102 $\pm$ 3 (16406) & 102 $\pm$ 3 (254)* & 101 $\pm$ 4 (625) & 101 $\pm$ 4 (716)** \\ 
Creatinine (mg/dL) & 1.0 $\pm$ 0.7 (16666) & 1.0 $\pm$ 0.7 (285)* & 0.9 $\pm$ 0.4 (648)** & 1.0 $\pm$ 0.6 (722)** \\ 
Glucose (mg/dL) & 109 $\pm$ 44 (16371) & 120 $\pm$ 46 (247)** & 126 $\pm$ 55 (606)** & 119 $\pm$ 51 (709)** \\ 
Potassium (mEq/L) & 4.4 $\pm$ 0.4 (16357) & 4.4 $\pm$ 0.5 (257) & 4.4 $\pm$ 0.5 (620) & 4.4 $\pm$ 0.6 (708)** \\ 
Sodium (mEq/L) & 140 $\pm$ 3 (16364) & 140 $\pm$ 3 (258) & 139 $\pm$ 4 (627)** & 140 $\pm$ 4 (713) \\ 
Total CO2 (mEq/L) & 24.0 $\pm$ 2.9 (16348) & 23.9 $\pm$ 3.2 (252) & 24.3 $\pm$ 3.2 (622)** & 24.1 $\pm$ 3.1 (708) \\ 
Total protein (g/dL) & 7.2 $\pm$ 0.5 (16805) & 6.9 $\pm$ 0.6 (303)** & 7.1 $\pm$ 0.7 (692)** & 7.0 $\pm$ 0.6 (721)** \\ 
\midrule
Basophils (cells/mL) & 0.0 $\pm$ 0.0 (16927) & 0.0 $\pm$ 0.0 (320)** & 0.0 $\pm$ 0.0 (660)** & 0.0 $\pm$ 0.0 (719)** \\ 
Basophils \% (\%) & 0.6 $\pm$ 0.4 (16944) & 0.5 $\pm$ 0.4 (325)** & 0.5 $\pm$ 0.4 (690)** & 0.5 $\pm$ 0.4 (728)** \\ 
Eosinophils (cells/mL) & 0.2 $\pm$ 0.2 (17162) & 0.2 $\pm$ 0.2 (359) & 0.2 $\pm$ 0.2 (700) & 0.2 $\pm$ 0.2 (728)* \\ 
Eosinophil \% (\%) & 2.8 $\pm$ 2.4 (17168) & 3.1 $\pm$ 3.0 (361) & 2.7 $\pm$ 2.7 (725)** & 2.6 $\pm$ 3.0 (739)** \\ 
Hematocrit (\%) & 40.9 $\pm$ 4.7 (16302) & 38.1 $\pm$ 5.8 (243)** & 37.7 $\pm$ 6.0 (621)** & 38.0 $\pm$ 5.7 (713)** \\ 
Hemoglobin (g/dL) & 13.5 $\pm$ 1.7 (16862) & 12.5 $\pm$ 2.1 (309)** & 12.5 $\pm$ 2.2 (675)** & 12.5 $\pm$ 2.0 (727)** \\ 
Hemoglobin A1c \% & 6.1 $\pm$ 1.2 (14112) & 6.4 $\pm$ 1.5 (140)** & 6.5 $\pm$ 1.5 (334)** & 6.4 $\pm$ 1.4 (540)** \\ 
Lymphocytes (cells/mL) & 2.0 $\pm$ 0.8 (16908) & 1.7 $\pm$ 0.8 (333)** & 1.7 $\pm$ 1.0 (663)** & 1.8 $\pm$ 1.4 (714)** \\ 
Lymphocytes \% (\%) & 31.7 $\pm$ 9.4 (16914) & 27.4 $\pm$ 11.7 (336)** & 26.0 $\pm$ 11.8 (691)** & 25.8 $\pm$ 12.2 (720)** \\ 
Neutrophils (cells/mL) & 3.7 $\pm$ 1.7 (16300) & 4.5 $\pm$ 3.5 (248)** & 4.3 $\pm$ 3.1 (577)** & 4.7 $\pm$ 3.1 (703)** \\ 
MCH (pg) & 29.7 $\pm$ 2.6 (16324) & 29.5 $\pm$ 3.3 (255) & 30.0 $\pm$ 3.0 (607)** & 29.9 $\pm$ 2.5 (707) \\ 
MCHC (g/dL) & 33.0 $\pm$ 1.2 (16341) & 33.0 $\pm$ 1.5 (259) & 33.0 $\pm$ 1.5 (589) & 32.8 $\pm$ 1.2 (708)** \\   
MCV (fL) & 89.8 $\pm$ 6.3 (16348) & 89.3 $\pm$ 7.7 (250) & 90.5 $\pm$ 7.0 (603) & 90.9 $\pm$ 6.2 (711)** \\ 
Monocytes (cells/mL) & 0.5 $\pm$ 0.2 (16220) & 0.6 $\pm$ 0.3 (255)** & 0.6 $\pm$ 0.3 (565)** & 0.6 $\pm$ 0.3 (702)** \\ 
Platelets (cells/mL) & 248 $\pm$ 72 (16829) & 256 $\pm$ 118 (314) & 223 $\pm$ 103 (679)** & 259 $\pm$ 110 (724) \\ 
RDW (\%) & 13.6 $\pm$ 1.4 (16329) & 14.5 $\pm$ 2.0 (242)** & 14.6 $\pm$ 2.0 (595)** & 14.5 $\pm$ 2.1 (715)** \\ 
WBC (cells/mL) & 6.5 $\pm$ 2.3 (17258) & 6.9 $\pm$ 3.6 (372) & 6.9 $\pm$ 3.5 (712) & 7.4 $\pm$ 3.8 (735) \\ 
\midrule
\multicolumn{5}{l}{* p-value $< 0.01$, ** p-value $ < 0.001$, based on Wilcoxon-rank sum tests comparing characteristics of cancer vs. control samples} \\
\bottomrule
\end{tabular}
\end{small}
\caption{Entire cohort characteristics of biomarkers (units) reported as mean and standard error (number of patients for whom the value was recorded). Biomarkers measured in either CMP or CBC are grouped accordingly. ALP, alkaline phosphatase; ALT, alanine aminotransferase; AST, aspartate aminotransferase; BUN, blood urea nitrogen; MCH, mean corpuscular hemoglobin; MCHC, mean corpuscular hemoglobin concentration; MCV, mean corpuscular volume; RDW, red cell distribution width; WBC, white blood cells.}
\label{tab:teststats_markers}
\end{center}
\end{table*}